\documentclass[12pt]{article}

\usepackage{caption}

\usepackage[T2A]{fontenc}
\usepackage[cp866]{inputenc}  

\usepackage{newlfont}
\usepackage{supertabular}
\usepackage{amsmath}
\usepackage{amssymb}
\usepackage{amsfonts}
\usepackage{amsthm}
\usepackage{bm}
\usepackage{mathtools}
\usepackage{pb-diagram}
\usepackage{righttag}
\usepackage{eqnarray}
\usepackage{subeqnarray}
\usepackage{algorithmic}
\usepackage{epsfig}
\usepackage{subfigure} 
\usepackage{emlines2}
\usepackage{rotating}
\usepackage{multirow}
\usepackage{graphicx}

\usepackage{tabularx}
\usepackage{parnotes}
\usepackage[sorting=none]{biblatex}
\usepackage{hyperref}
\usepackage[english]{babel}

\DeclareFieldFormat{labelnumber}{#1} 

\usepackage{color}

\usepackage{booktabs}
\usepackage{url}
\usepackage{float}

\usepackage{fancybox}

\theoremstyle{plain}

\theoremstyle{definition}
\newtheorem{example}{Пример}

\newtheoremstyle{break}
  {9pt}
  {9pt}
  {\itshape}
  {}
  {\bfseries}
  {.}
  {\newline}
  {}

\theoremstyle{break}

\makeatletter

\renewcommand{\section}{\@startsection{section}{1}%
             {\parindent}{3.5ex plus 1ex minus .2ex}%
             {2.3ex plus.2ex}{\normalsize\bf}}

\renewcommand{\subsection}{\@startsection{subsection}{2}%
             {\parindent}{3.5ex plus 1ex minus .2ex}%
             {2.3ex plus.2ex}{\normalsize\bf}}

\renewcommand{\subsubsection}{\@startsection{subsubsection}{3}%
             {\parindent}{3.5ex plus 1ex minus .2ex}%
             {2.3ex plus.2ex}{\normalsize\bf}}

\renewcommand{\paragraph}{\@startsection{paragraph}{4}%
             {\parindent}{3.5ex plus 1ex minus .2ex}%
             {2.3ex plus.2ex}{\normalsize\bf}}

\renewcommand{\@biblabel}[1]{#1\hfill}

\makeatother

\graphicspath{{Picts/}}

\newcounter{partnumber}

\makeatletter
\@addtoreset{equation}{section}
\renewcommand{\theequation}{\arabic{equation}}

\@addtoreset{table}{section}

\makeatother

\newcounter{fragm}
\newcounter{subfragm}[fragm]

\newcounter{myremark}[section]

\newcounter{myalgorithm}[section]

\usepackage{float}
\usepackage{tabularx}

\usepackage{xcolor}
\usepackage{nicematrix}

\newcolumntype{C}{>{\centering\arraybackslash}X}  
\newcolumntype{P}[1]{>{\centering\arraybackslash}p{#1}} 

\usepackage{caption}
\newlength{\DepthReference}
\newlength{\HeightReference}
\newlength{\Width}%

\newcommand{\MyColorBox}[2][red]%
{%
    \settowidth{\Width}{#2}%
    \colorbox{#1}%
    {%
        \raisebox{-\DepthReference}%
        {%
                \parbox[b][\HeightReference+\DepthReference][c]{\Width}{\centering#2}%
        }%
    }%
}
\newcommand{\MyTBox}[2]%
{%
    \settowidth{\Width}{#1}%
    \fbox%
    {%
        \raisebox{-\DepthReference}%
        {%
                \parbox[b][\HeightReference+\DepthReference][c]{\Width}{\centering#1}%
        }%
    }%
}

\usepackage{xcolor}

\begin{document}

\begin{center}
\textbf{Open-Vocabulary Indoor Object Grounding \\ with 3D Hierarchical Scene Graph}
\end{center}

\begin{center}
Linok Sergey$^{1*}$, Naumov Gleb$^{1}$\\
$^{1}$Moscow Institute of Physics and Technology, Moscow, Russia \\
$^{*}$e-mail: linok.sa@phystech.edu \\
\end{center}

\textbf{Abstract.}
We propose \textbf{OVIGo-3DHSG} method - \textbf{O}pen-\textbf{V}ocabulary \textbf{I}ndoor \textbf{G}rounding of \textbf{o}bjects using \textbf{3D} \textbf{H}ierarchical \textbf{S}cene \textbf{G}raph. OVIGo-3DHSG represents an extensive indoor environment over a Hierarchical Scene Graph derived from sequences of RGB-D frames utilizing a set of open-vocabulary foundation models and sensor data processing. The hierarchical representation explicitly models spatial relations across floors, rooms, locations, and objects. To effectively address complex queries involving spatial reference to other objects, we integrate the hierarchical scene graph with a Large Language Model for multistep reasoning. This integration leverages inter-layer (e.g., room-to-object) and intra-layer (e.g., object-to-object) connections, enhancing spatial contextual understanding. We investigate the semantic and geometry accuracy of hierarchical representation on Habitat Matterport 3D Semantic multi-floor scenes.  Our approach demonstrates efficient scene comprehension and robust object grounding compared to existing methods. Overall OVIGo-3DHSG demonstrates strong potential for applications requiring spatial reasoning and understanding of indoor environments. Related materials can be found at \url{https://github.com/linukc/OVIGo-3DHSG}.\footnote[1]{This work has been submitted to the OMNN for publication. Copyright may be transferred without notice, after which this version may no longer be accessible.}


\bigskip
\section{Introduction}
\label{sec:intro}

Understanding and interpreting indoor environments is essential for numerous practical applications~\cite{mironov2023strl,linok2023influence}, such as robotics, augmented reality, smart home systems, and assistive technologies. Classical methods for representing the geometry of the surrounding space include SLAM~\cite{macario2022comprehensive} approaches and their more modern interpretations, such as NeRF~\cite{mildenhall2021nerf} and Gaussing Splatting~\cite{kerbl20233d}, while various neural network models for detection and segmentation are actively employed for object recognition~\cite{lee2024comparative,minaee2021image}. Notably, the latter~\cite{zou2023segment,liu2024grounding,cheng2024yolo} can operate in an open-vocabulary mode, potentially recognizing a larger number of classes and doing so with flexible arbitrary input. Modern approaches~\cite{sun2024sparse,zou20253d,yu2025rgb,yudin2024multimodal} that integrate both object and geometric representations, enabling the encoding of textual descriptions into the scene geometry for subsequent retrieval based on natural language queries, are gaining popularity also.

However, as scene sizes increase, the complexity of object search becomes a significant challenge. To address this issue, methods that structure the recognized object representations into a graphical hierarchy are being developed. In indoor environments, the primary levels of such hierarchy are typically floors and rooms. Degree to which the search space can be effectively narrowed, according to the user's query and thus the resource efficiency of entity search within large scenes, is largely dependent on the topology of the hierarchy and the reasoning approach over resulted hierarchical graph.

\begin{figure}[H]
  \centering
  \includegraphics[width=1.0\textwidth]{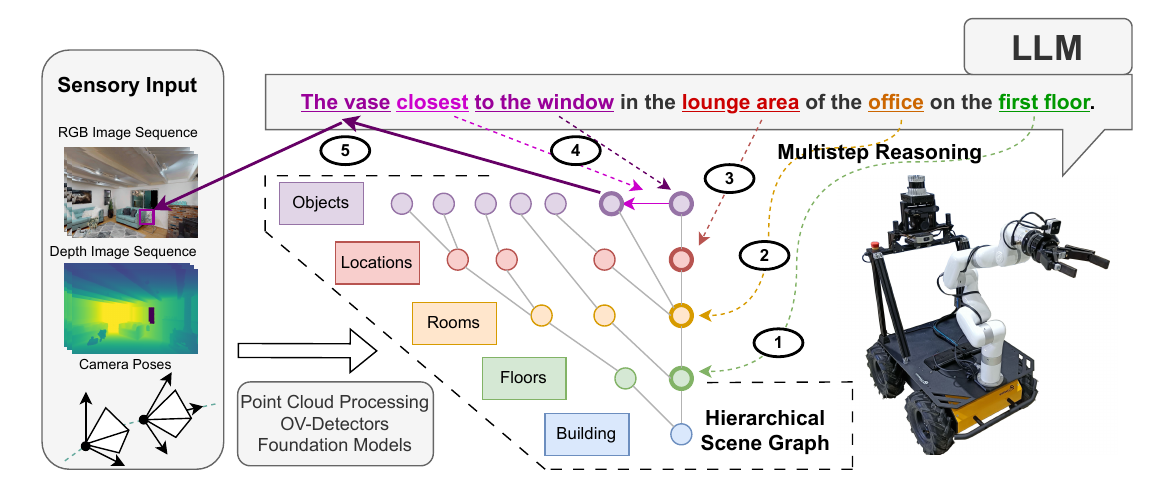}
  \caption{We represent vast observed areas based on a sequence of RGB-D frames in the form of a Hierarchical Scene Graph using a set of open-vocabulary foundation models and sensor data processing. For the indoor environment, this hierarchy includes floors, rooms, locations, and objects. To handle complex spatial object queries we integrate Hierarchical Scene Graph into a Large Language Model for multistep reasoning. Utilizing both inter-layer and intra-layer object connections allows efficient and robust scene understanding and 3D object grounding in a free form of natural language.}
  \label{fig:graphical_abstract}
\end{figure}

In this work we present \textbf{OVIGo-3DHSG} - \textbf{O}pen-\textbf{V}ocabulary \textbf{I}ndoor \textbf{G}rounding of objects with a help of \textbf{3D} \textbf{H}ierarchical \textbf{S}cene \textbf{G}raph, that represents extensive indoor environment using Hierarchical Scene Graph derived from sequences of RGB-D frames using a set of open-vocabulary foundation models and sensor data processing. The hierarchical representation explicitly models spatial relations across floors, rooms, locations, and objects. To effectively address complex object queries, we integrate the hierarchical scene graph with a Large Language Model for multi-step reasoning. This integration leverages inter-layer (e.g., room-to-object) and intra-layer (e.g., object-to-object) connections, enhancing contextual understanding. You can find the concept illustration in Fig.~\ref{fig:graphical_abstract}.

In summary, we make the following contributions:
\begin{itemize}
  \item new ``location'' level of hierarchy in an indoor environment to further structure grouped objects belonging to a meaningful area inside room (Sec.~\ref{method:location}, Sec.~\ref{exp:locdetector});
  \item robust time-efficient open-vocabulary object layer construction method using a set of foundation models (Sec.~\ref{method:object});
  \item multistep LLM reasoning algorithm for object grounding based on 3DHSG that not only utilizes intra-layer connections but also inter-layer object relations to answer complex spatial text quires (Sec.~\ref{method:inference}, Sec.~\ref{exp:grounding});
  \item evaluations on ScanNet\texttt{+}\texttt{+}v2~\cite{yeshwanthliu2023scannetpp} and HM3DSem~\cite{yadav2023habitat} datasets that include semi-auto labeled location masks and question-answer pairs for benchmarking complex natural language quires about hierarchical environments (Sec.~\ref{locdetector:dataset}, Sec~\ref{grounding:dataset}).
\end{itemize}

We make related materials available at this url\footnote{\url{https://github.com/linukc/OVIGo-3DHSG}}.

\section{Related Work}
\label{sec:related_works}
\subsection{Hierarchical Indoor 3D Scene Graph Construction}

3DSG~\cite{armeni20193d} is one of the first approaches that present a hierarchical representation of indoor environments. This method proposes a four-layered graph: building, rooms, objects, and cameras, which serves as a navigation layer, i.e., the sensor's (camera's) trajectory. 3DDSG~\cite{rosinol20203d} extends the functionality of the graph by incorporating the trajectories (nodes and edges) of moving objects. Hydra~\cite{hughes2022hydra}, S-Graphs$+$~\cite{bavle2023s}, and Foundations~\cite{hughes2024foundations} combine the processes of geometric perception (localization) and constructing an object hierarchical graph representation, which actively uses a combination of point cloud processing and other heuristics, into a unified approach.  Such methods are considered to be classical now and are actively investigated in current research. Modern HOV-SG~\cite{werby2024hierarchical} method integrates foundation CLIP features into a single pipeline, allowing closed-vocabulary object detectors to be substituted with an open-vocabulary 3D scene understanding approach.

In our work OVIGo-3DHSG we extend the levels of abstraction in the graph by adding ``locations'' that represent a geometrically related group of objects within rooms, helping reduce searching space, which is crucial for a vast indoor scenes object search and also adds another layer of abstaction for a user's query.

\subsection{Applying Scene Graph for 3D Object Grounding}

Using a constructed structured graph representation for object grounding based on user queries, especially with spatial relations to other objects, is an effective method due to the compact diverse textual (and other) node and edge properties compression. For example, in 3DSG~\cite{armeni20193d} a trainable neural network-based algorithm is presented, which predicts spatial relationships between object masks in an image. SceneVerse~\cite{jia2024sceneverse} entirely relies on a set of rules based on the intersections of 3D bounding boxes, which limits its robustness. ConceptGraphs~\cite{gu2024conceptgraphs} employs a large language model to determine a finite set of relationships based on the 3D coordinates of objects, which significantly reduces the descriptive capacity of such a graph. Recent approaches tend to use a mixture of heuristic algorithms for determining spatial relationships along with a pre-trained language model to reason regarding the graph's data, as presented by the BBQ~\cite{linok2024beyond}, where in addition to the relationships from ZSVG3D~\cite{yuan2024visual} metric edges are also computed. 3DGraphLLM~\cite{zemskova20243dgraphllm} is a method for constructing a learnable representation of a 3D scene graph, which serves as input for LLMs to perform 3D vision-language tasks.

Modern approaches~\cite{tang2024openobject,maggio2024clio} are also actively investigating hierarchical graph utilization to facilitate reasoning and performing low-level tasks (object grounding, question answering, planning, navigation, etc.) by user textual instructions. However, a common pattern is that they do not use intra-layer object relations which may be crucial when we ground objects based on relations to others. Also, the mere presence of relations does not enable the graph to be used effectively for search. Constructing edges between all objects is not only resource-intensive but also significantly increases the context supplied to the language model, which may lead to hallucinations in language model responses. 

Therefore, in OVIGo-3DHSG, we propose an adaptation of the deductive algorithm from BBQ~\cite{linok2024beyond} for multistep reasoning with a hierarchical representation. Our approach uses a large language model to first select the objects relevant to the user's query and then build the connections in the graph, utilizing this information to perform an object grounding based on a textual query with valuable intra-layer and inter-layer connections.

\section{Method}
\label{sec:method}

\begin{figure}
    \centering
    \includegraphics[width=1.0\linewidth]{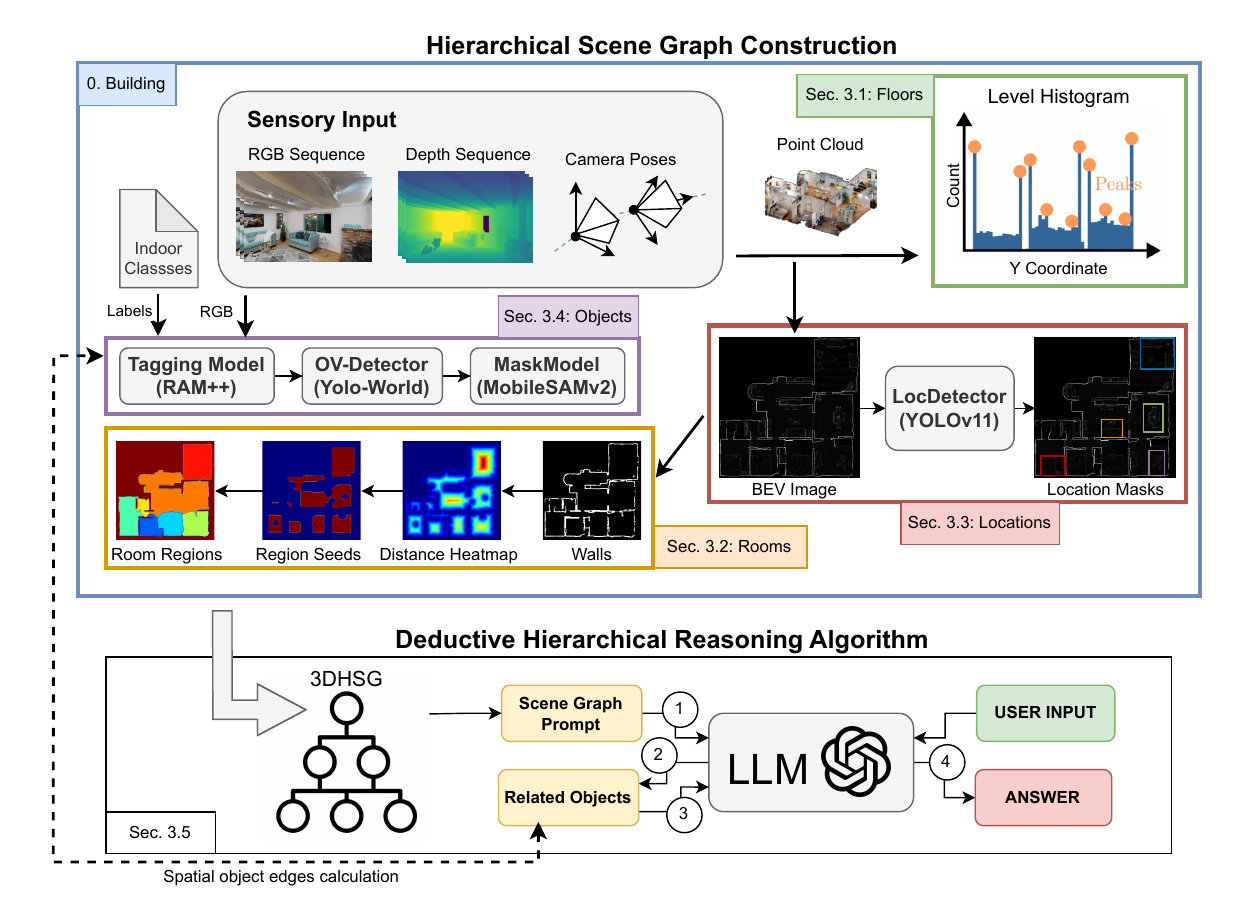}
    \caption{Pipeline of OVIGo-3DHSG: Hierarchical Scene Graph construction based on Sensory Input and Deductive Hierarchical Reasoning Algorithm to inference user's text query.}
    \label{fig:method}
\end{figure}

Given a sequence of posed RGB-D frames, OVIGo-3DHSG constructs a hierarchical scene graph (HSG) that consists of five layers. We formalize our graph as $G = (N, E)$, where $N$ denotes the nodes and $E$ denotes the edges. The nodes can be expressed as $N = N_B \cup N_F \cup N_R \cup N_L\cup N_O$, consisting of a root node (building node) $N_B$ , floor nodes $N_F$ (Sec.~\ref{method:floor}), room nodes $N_R$ ((Sec.~\ref{method:room})), location nodes $N_L$ (Sec.~\ref{method:location}) and object nodes $N_O$ (Sec.~\ref{method:object}). Each node in the graph except the root node $N_B$ contains the point cloud, bounding box and mask of the concept it refers to and text tag. Relations in the graph represent both inter $E_{inter}$ and intra $E_{intra}$ layer connections and can be written as $E = E_{inter} \cup E_{intra}$. Here, $E_{inter} = E_{BF} \cup E_{FR} \cup E_{RL} \cup E_{RO} \cup E_{LO}$ represents the boolean edges that are obtained after sensory input perception: $E_{BF}$ is an edge between the root node and the floor nodes, $E_{FR}$ represents the edges between the floor nodes and the room nodes, $E_{RL}$ represents the edges between the room nodes and the location nodes and lastly, $E_{RO}$ and $E_{LO}$ denotes the edges between the room or the location and object nodes correspondently. $ e_{inter} \in E_{inter}$ describes existance of connection between each element based on spatial overlapping ($e_{inter} \in \{0, 1\}$). $E_{intra} = E_{metric} \cup E_{semantic}$ and represents spatial edges between objects. This edges are computed for only related objects to the input user's text query based on deductive hierarchical reasoning algorithm (Sec.~\ref{method:inference}). Overall pipeline of OVIGo-3DHSG is illustrated in Fig.~\ref{fig:method}.

\subsection{Floor Layer}
\label{method:floor}

For floor segmentation, we build a height histogram of all points from the point cloud $P$, accumulated from the posed RGB-D sequence, with a bin size of $bin_h$ meters. In this histogram, we identify peaks using a neighborhood size of $\delta_{f}$ meters. Let $h_{max}$ denote the highest value among the detected peaks. We retain only the peaks with heights $h$ such that $h > p_hh_{max}$, where $p_h \in (0,1)$. We then apply DBSCAN~\cite{DBSCAN} to the histogram and, in each cluster, we keep the two peaks with the highest values, which will serve as the boundaries of a floor. This approach effectively reflects the dataset creation process, namely that during point cloud scanning there is free space between the ceiling of floor $i$ and the floor of floor $i+1$. Afterward, every two peaks in the sorted list of the remaining peaks form the boundaries of a floor. Each floor also gets a tag of type ``floor i''. For the experiments, we set $bin_h = 0.01, \delta_f=0.2, p_h=0.9$.

\subsection{Room Layer}
\label{method:room}

Room segmentation is performed separately on each floor. Let $P_{Fi}$ denote the point cloud of floor $i$. It is projected onto the horizontal plane, resulting in a BEV image $BEV_{Fi} \in \mathbb{R}^{H \times W}$. Additionally, min-max normalization is applied so that the pixel values lie in the range $[0, 1]$. From the $BEV_{Fi}$, a binary mask \(M_{wall}\) is generated using a threshold $\delta_{wall}$, thereby yielding a binary map of tall objects such as walls. Then, based on this wall map, a Euclidean Distance Field ${EDF}$ is constructed, where each pixel is assigned a value equal to the minimum Euclidean distance to a pixel with a positive value in $M_{wall}$. From the $EDF$, a binary Region Seeds mask is created using Otsu's method~\cite{Otsu}. Finally, the Watershed algorithm~\cite{kornilov2018overview} is applied to obtain the final room masks. To assign a text tag to the each room we prompt LLM with a list of object and location tags it contains with a task of identifying the room type.

\subsection{Location Layer}
\label{method:location}

We define a location as a portion of the room that contains objects that are close enough to one another and share common semantic properties. Locations may correspond to groups of objects such as a ``bed with bedside tables'', ``a sofa with a pouf in front of it'' or a ``kitchen area inside an openspace''. For each floor, we project $P_{Fi}$ onto the horizontal plane to obtain a $BEV_{Fi} \in \mathbb{R}^{H \times W}$ as for room segmentation, but during the projection we retain only those points $p \in P_{Fi}$ that satisfy
$
b_{min} + \alpha_{min}(b_{max} - b_{min}) \leq \mathbf{p}_z \leq b_{min} + \alpha_{max}(b_{max} - b_{min}),
$
criteria, where 
$
b_{min} = \min_{\mathbf{p} \in \mathbb{P}} \mathbf{p}_z, \quad b_{max} = \max_{\mathbf{p} \in \mathbb{P}} \mathbf{p}_z,
$
and $\alpha_{min}$ and $\alpha_{max}$ are hyperparameters. In this manner, we eliminate objects hanging from the ceiling, such as ``chandeliers'', as well as noise at the floor level. The resulting $BEV_{Fi}$ is then fed as input to a trained neural network-based LocDetector, which outputs location masks. More details about LocDetector realization can be found in Sec.~\ref{exp:locdetector}. To assign a text tag to the location we prompt LLM with a list of object tags it contains with a task of identifying the location type.

\subsection{Object Layer}
\label{method:object}

For each input RGB-D frame of image $I_{i}$, depth image $D_{i}$ and pose $\mathcal{P}_{i}$ we first extract text tags from $I$ using tagging model RAM\texttt{+}\texttt{+}~\cite{zhang2023recognize} from a large corpus over a hundred possible indoor labels. Next, we consecutively call open-vocabulary detector Yolo-World~\cite{cheng2024yolo} with a tag prompt to identify 2D bounding boxes on the image $I$ and MobileSAMv2~\cite{zhang2023mobilesamv2} mask segmentation model with a prompt of 2D bounding boxes to segment mask for each object. Detections on each frame from the sequence are projected in the 3D space using $D_{i}$, $\mathcal{P}_{i}$ and sensor parameters. At the end of the sequence, we aggregate objects with spatial and semantic constraints, effectively reducing duplicates from multiple object observations, assigning for each object multiview tags.

\subsection{Deductive Hierarchical Reasoning Alghorithm}
\label{method:inference}

To prompt a large language model with scene graph information we develop Deductive Hierarchical Reasoning algorithm. Given scene graph $G = (N, E_{inter})$ after the Hierarchical Scene Graph Construction step and user text query $Q$, we apply multistep reasoning approach that consists of multiple LLM-based calls to select related objects over multiple levels of hierarchy (Sec.~\ref{sec:related}) and grounding step that also use LLM to find unique identifier of required object (Sec.~\ref{sec:grounding}).

\subsubsection{Selecting Related Objects}
\label{sec:related}

In the first step, we try to reduce search space and retrieve subgraphs that relate to the user's query $Q$. For each level of the hierarchy, we prompt input text request and corresponding layer entity tags with unique IDs to choose only related entities. As a result, we get $N$ related subgraphs of $G$. In each subgraph's object group, we ask LLM to select target IDs and anchor IDs. Anchors represent objects to which LLM needs to know relations to answer user query $Q$. Then we enrich selected textualized object nodes inside each group with semantic $E_{semantic}$ and metric edges $E_{metric}$ between targets and anchors. $E_{semantic}$ are calculated similar to ZSVG3D~\cite{yuan2024visual} resulting in edges like ``left'', ``right'', ``back'', ``front'', ``above'' and ``below''. $E_{metric}$ are calculated by comparing 3D object boxes in a template like ``object A with id 1 is in 3 meters of object B with id 2''. As a result of related objects selection we have $N$ groups of object-layer graphs $\hat{G}=(\hat{N}_{O}, E_{intra})$, where $\hat{N}_{O}$ - choosed target and anchor objects.

\subsubsection{Object Grounding}
\label{sec:grounding}

We prompt input text request $Q$ and corresponding textualized $\hat{G}$ in LLM to perform the task of grounding object based on provided unique identifiers associated within each object. Using the compact scene description of relevant objects, LLM retrieves the final object in JSON format. To enhance the performance of large language models on complex reasoning tasks we also incorporate into prompt examples, such as intermediate and output formats; reasoning and explanation fields that guide the model through a chain-of-thought~\cite{wei2022chain} and allow the LLM to break down multistep reasoning problems into more manageable parts, leading to better accuracy and interpretability. If the response needs formatting, we make another LLM call to adjust it to later parse output data in JSON.

\section{Experiments}

In the experiments section we provide a detailed description of the training and inference results of LocDetector (Sec.~\ref{exp:locdetector}), and analyze the performance of the OVIGo-3DHSG on the object grounding task both on the open evaluation data and created benchmark of complex hierarchical text queries (Sec.~\ref{exp:grounding}). Our primary data source for evaluation are 8 multi floor scenes from HM3DSem~\cite{yadav2023habitat}: \textit{00824}, \textit{00829}, \textit{00843}, \textit{00861}, \textit{00862}, \textit{00873}, \textit{00877}, \textit{00890}). We choose the same scenes as from HOV-SG~\cite{werby2024hierarchical} to be aligned with previous research.

\subsection{Location Detector Module}
\label{exp:locdetector}

\subsubsection{Dataset Labeling}
\label{locdetector:dataset}

For the training location detection model LocDetector, we used the Scannet++v2 dataset~\cite{yeshwanthliu2023scannetpp}, which contains 1006 3D scenes overall. Scenes that either did not allow locations to be defined according to our criteria or where the entire scene constituted a location were excluded. Our final dataset comprised 546 scenes, with 400 scenes selected for the training set and 146 for the validation set. In each scene, the maximum number of locations did not exceed 15. The annotation of each location is a 2D contour (mask) on a BEV image.

For scene annotations, we applied a hybrid approach consisting of a clustering algorithm followed by manual annotation refinement and validation of the obtained results. Let a scene be represented as a point cloud (Eq.~\ref{eq1})
\begin{equation}
\label{eq1}
P = \{\mathbf{p}_1, \mathbf{p}_2, \dots, \mathbf{p}_n\}, \quad \mathbf{p}_i \in \mathbb{R}^3,
\end{equation}
and assume that each point belongs to some object (Eq.~\ref{eq2}), i.e., 
\begin{equation}
\label{eq2}
P = \bigsqcup\limits_{i=1}^l O_i.
\end{equation}
Let minimum and maximum height points (Eq.~\ref{eq3}) be
\begin{equation}
\label{eq3}
b_{min} = \min\limits_{\mathbf{p} \in P} \mathbf{p_z},\quad b_{max} = \max\limits_{\mathbf{p} \in P} \mathbf{p_z}.
\end{equation}
The new point cloud is obtained by removing the high and low points (Eq.~\ref{eq4}):

\begin{equation}
\label{eq4}
\hat{P} = \left\{\mathbf{p} \in \mathcal{P} \mid b_{min} + \alpha_{min}\left(b_{max} - b_{min}\right) \leq \mathbf{p_z} \leq b_{min} + \alpha_{max}\left(b_{max} - b_{min}\right)\right\},
\end{equation}
where \(\alpha_{min}, \alpha_{max}\) are hyperparameters. The DBSCAN algorithm~\cite{DBSCAN} is then run on \(\hat{P}\).

Consider the clustering output (Eq.~\ref{eq5})
\begin{equation}
\label{eq5}
\mathcal{C} = \{ C_1, C_2, \dots, C_k \}, \quad C_j \subset \hat{P},
\end{equation}
and apply the following heuristics to these clusters:
\begin{enumerate}
    \item For each object \(O\), if there exist non-noise points within it, then all points of the object are assigned to the cluster that contains the largest number of its points.
    \item If a cluster contains fewer than \(c\) objects, it is discarded.
\end{enumerate}

The resulting clusters are projected onto the horizontal plane (Eq.~\ref{eq6}):
\begin{equation}
\label{eq6}
\pi : \mathbb{R}^3 \to \mathbb{R}^2, \quad \pi(x, y, z) = (x, y), \quad \pi(C_j) = \{ \pi(\mathbf{p}) \mid \mathbf{p} \in C_j \} \subset \mathbb{R}^2.
\end{equation}
Alphashape is then applied to the projected clusters $\pi(C_j)$. Polygons with small areas, as well as those that are overly elongated, are removed (Eq.~\ref{eq7}). Specifically, if \(PR\) is the perimeter of the polygon describing a cluster and \(S\) is its area, then polygons satisfying
\begin{equation}
\label{eq7}
\frac{4 \pi S}{PR^2} < C
\end{equation}
are discarded.

\subsubsection{Training}
\label{locdetector:training}

Experiments were conducted with two models: RoomFormer~\cite{yue2023connecting} and YOLOv11~\cite{khanam2024yolov11}. We use RoomFormer without changes in its architecture. In the output layer, we employed the following dimensions: 
$num\_polys=15,\, num\_queries=570$, where $num\_polys$ is the maximum number of predicted polygons (locations), and $num\_queries$ is the maximum number of points per polygon (location). 
We pre-trained the model for 20 epochs on the Structured3D dataset~\cite{Structured3D} for room detection using the original RoomFormer hyperparameters and then fine-tuning on our annotated Scannet++v2 dataset~\cite{yeshwanthliu2023scannetpp}. We use AdamW optimizer with a $weight\_decay\_factor=1e\text{-}4$ and $batch\ size=16$. We train the model for 400 epochs with a learning rate of $5e\text{-}5$.  We train YOLOv11 on the instance segmentation task with a single class (location) for 300 epochs using the AdamW optimizer with parameters $learning\_rate=0.002$, $\beta_1,\beta_2=0.9, 0.999$. Both models were trained on an Nvidia RTX3060 Ti with 8 Gb VRAM.

\subsubsection{Metrics}
\label{locdetector:evaluation}

Let $\delta \in (0, 1)$ be a fixed threshold, and let 
$\mathcal{M} = \{M_1, \dots, M_k\}$ and $\mathcal{M}^{gt} = \{M^{gt}_1, \dots, M^{gt}_l\}$ 
denote the predicted 2D masks and the ground truth masks of the locations, respectively. 
For each predicted mask $M_i \in \mathcal{M}$, the IoU is computed with all ground truth masks in $\mathcal{M}^{gt}$:
$I_{i,j} = \text{IoU}\left(M_i, M_j^{gt}\right)$. Let $k = \arg\max_j I_{i,j}$. 
Then, if $I_{i,j} > \delta$, the prediction $M_i$ is considered a TP and the ground truth mask $M_j$ is removed 
from the set $\mathcal{M}^{gt}$; otherwise, the prediction $M_i$ is considered a FP. 
After all $M_i \in \mathcal{M}$ have been processed, the unmatched elements in $\mathcal{M}^{gt}$ are treated as FNs. 
These values allow the computation of the F-measure, denoted as $F_1@\delta$.

\begin{figure}[t]
    \centering
    \includegraphics[width=0.75\linewidth]{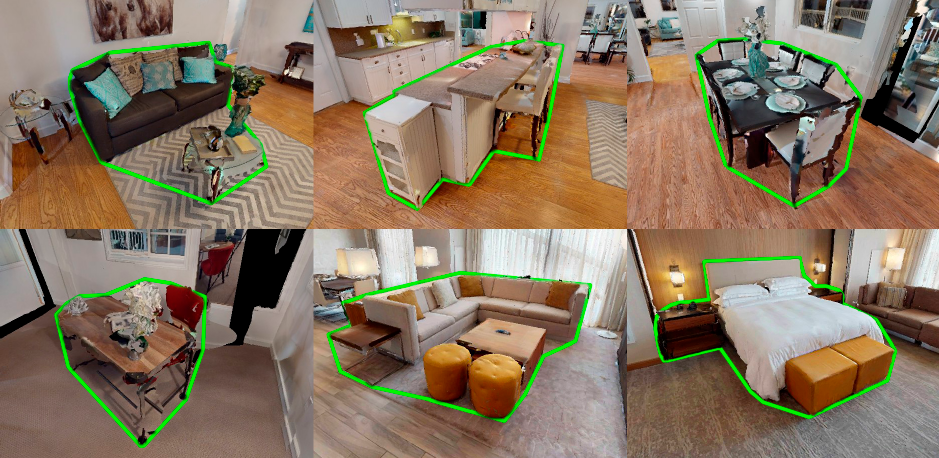}
    \caption{Qualitative example of YOLOv11~\cite{yolo11_ultralytics} LocDetector predictions on HM3DSem~\cite{yadav2023habitat}. The green outline in the image is the result of location detection post-processing, namely the outline aggregation of objects that fall entirely within the predicted location mask.}
    \label{fig:quality_loc}
\end{figure}

\begin{figure}[t]
    \centering
    \begin{minipage}{0.5\textwidth}
       \centering
       \includegraphics[width=\textwidth]{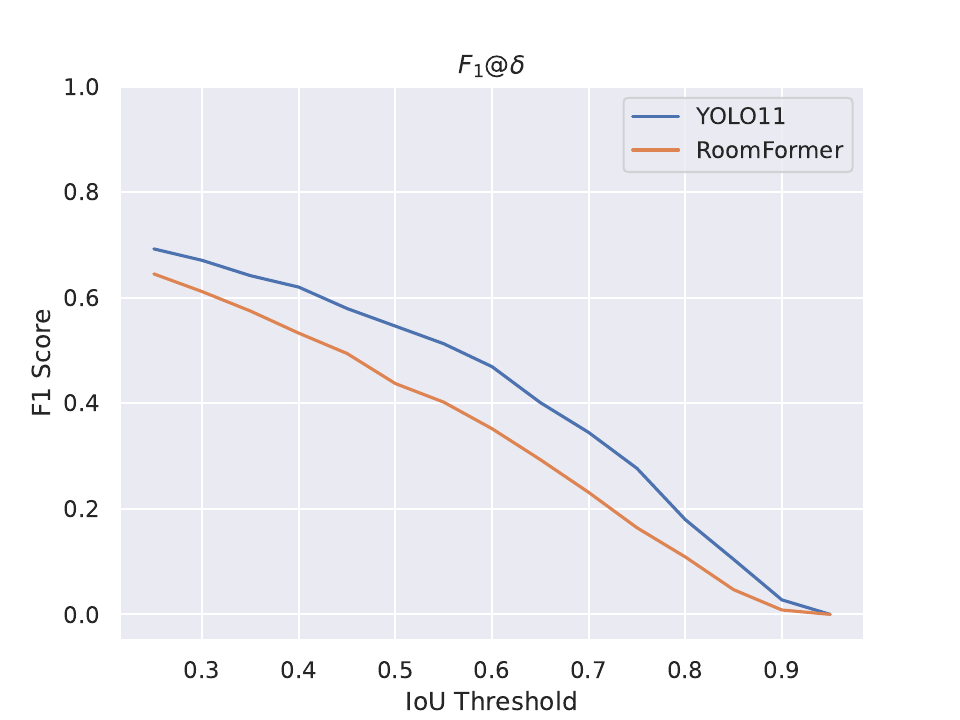}
       \caption{Validation quantitative results of location detection models on ScanNet\texttt{+}\texttt{+}v2~\cite{yeshwanthliu2023scannetpp}.}
       \label{fig:f_scannet}
    \end{minipage}\hfill
    \begin{minipage}{0.5\textwidth}
       \centering
       \includegraphics[width=\textwidth]{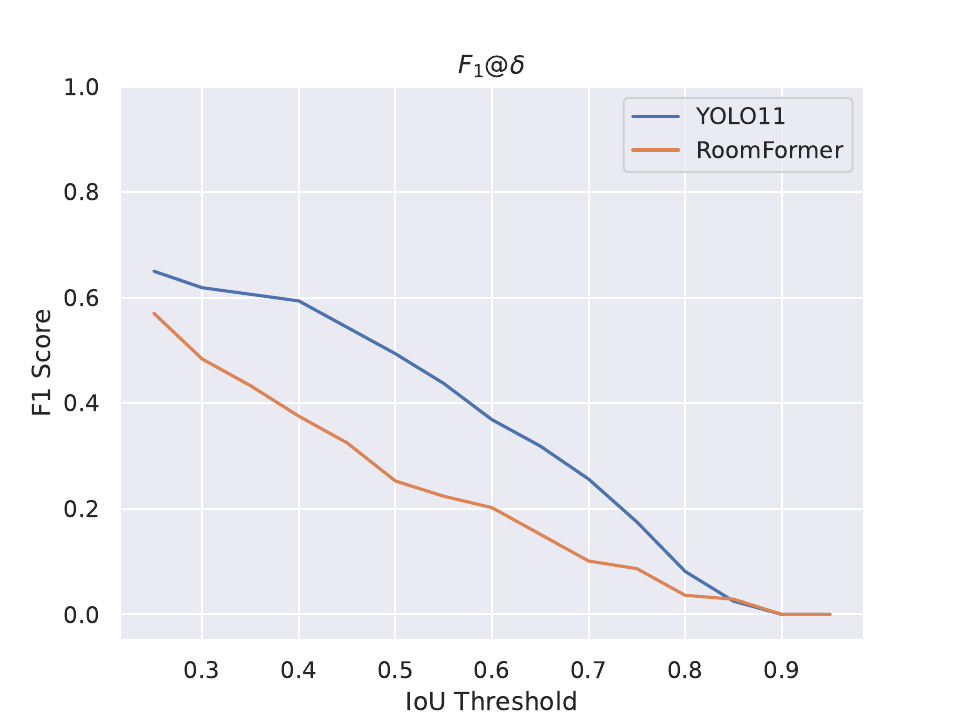}
       \caption{Quantitative results of location detection models on 8 labled scenes from HM3DSem~\cite{yadav2023habitat}.}
       \label{fig:f_habitat}
    \end{minipage}
\end{figure}

\subsubsection{Results}
\label{locdetector:comments}

Fig.~\ref{fig:f_scannet} shows quantitative results for both models on validation split of our location dataset on ScanNet\texttt{+}\texttt{+}v2. In the presented graphic, it is evident that the YOLOv11 model performs better at solving the task. A similar trend is observed when transferring to another data source. Fig.~\ref{fig:f_habitat} shows quantitative results for both models on 8 labeled scenes from HM3DSem~\cite{yadav2023habitat}. A possible reason for such behavior may be a combination of two factors: a simpler formulation of the task for the YOLOv11 model (binary segmentation) compared to predicting contours using the transformer-based RoomFormer model, and the size and diversity of the training dataset for the models. For further improving the quality of location detection models, the latter factor seems most critical, especially given the lack of open datasets on this topic. Fig.~\ref{fig:quality_loc} shows qualitative example of YOLOv11 LocDetector predictions on HM3DSem~\cite{yadav2023habitat}.

\subsection{Object Grounding Task}
\label{exp:grounding}

\subsubsection{Dataset}
\label{grounding:dataset}

To highlight the importance of intra-layer object connections answering complex text queries, we create our 3D object grounding benchmark on the same as mentioned before 8 scenes from HM3DSem~\cite{yadav2023habitat}. A distinctive feature of this benchmark is that the question contains references to other objects, without which it is impossible to unambiguously determine which target object is being referred to in the answer, like "find a vase near the window" or "find the leftmost pillow on the sofa in the living room". The provided annotation consists of 50 questions, with a 3D bounding box defined as the corresponding answer for each query.

\subsubsection{Metrics}
\label{grounding:evaluation}

In our work for object-level semantic evaluation, we reuse $AUC^{k}_{top}$ metric from HOV-SG~\cite{werby2024hierarchical} that quantifies the area under the $top_k$ accuracy curve between the predicted and the actual ground-truth object category for comparison evaluation. Metric encodes how many erroneous shots are necessary on average before the ground-truth label is predicted correctly. Based on this, the
metric encodes the actual open-set similarity while scaling to large, variably-sized label sets. For object grounding task we use Acc@0.1, Acc@0.25, Acc@0.5, and Acc@0.75 metrics. Prediction is considered to be true positive if the Intersection over Union (IoU) between the selected object bounding box and the ground truth bounding box exceeds 0.1, 0.25, 0.5, and 0.75, respectively.

\subsubsection{Results}
\label{grounding:comments}

To demonstrate the advantages of the proposed approach, we first experiment with an object-level representation. Our approach directly assigns labels to the objects, therefore, to compare object-level representation with previous research from VLMaps~\cite{huang2023visual}, ConceptGraphs~\cite{gu2024conceptgraphs}, HOV-SG~\cite{werby2024hierarchical} that use ranking based on implicit CLIP latent image representation, we utilize text-based similarity ranking between all classes and assigned per-object multiview tags. With this experiment (shown in Tab.~\ref{tab:hovsg_objects}), we want to highlight that the proposed approach generates a sufficient number of objects and better covers the observed sensory data (in terms of correct class), even though it is based on a drastically different detection principle and generates sparse object observations compare to ``segment and encode/describe all strategy".

To demonstrate the importance of edges between objects for 3D object grounding based on complex spatial queries, we compare with HOV-SG~\cite{werby2024hierarchical}, which is the closest baseline compared to our method, but does not have connections between objects. In this experiment, our approach predicts object ID as described in Sec.~\ref{method:inference}. We use ChatGPT4o-mini~\cite{hurst2024gpt} as a primarily large language model. The results in Tab.~\ref{tab:habitat_objects} show that for successful grounding based on complex quires, an idea from our approach to adding spatial relations between objects plays a key role. In addition, it is worth noting that these edges between objects are not built by exhaustive search, but with a restriction only on relevant objects, which does not add significant costs in terms of computing resources.

\begin{table}[t]
    \centering
    \caption{Quantitative results of object-level semantics evaluation on HOV-SG~\cite{werby2024hierarchical} annotations on HM3DSem~\cite{yadav2023habitat}.}
    \label{tab:hovsg_objects}
    \begin{tabularx}{\textwidth}{|p{4.cm}|p{1.3cm}|p{1.3cm}|p{1.3cm}|p{1.3cm}|p{1.3cm}|p{1.3cm}|p{1.7cm}|}
        \toprule 
        Method & $top_{5}$ & $top_{10}$ & $top_{25}$ & $top_{100}$ & $top_{250}$ & $top_{500}$ & $AUC^{k}_{top}$\\
        \midrule
        VLMaps~\cite{huang2023visual} & 0.05 & 0.17 & 0.54 & 15.32 & 26.01 & 40.02 & 56.20 \\
        ConceptGraphs~\cite{gu2024conceptgraphs} & 18.11 & 24.01 & 33.00 & 55.17 & \textbf{70.85} & \textbf{81.55} & 84.07 \\
        HOV-SG~\cite{werby2024hierarchical} & 18.43 & 25.73 & 36.41 & \textbf{56.46} & 69.95 & 80.86 & 84.88 \\
        \midrule
        OVIGo-3DHSG (ours) & \textbf{19.14} & \textbf{26.61} & \textbf{40.16} & 54.23 & 68.45 & 79.84 & \textbf{85.13} \\
        \bottomrule
    \end{tabularx}
\end{table}

\begin{table}[t]
    \centering
    \caption{Quantitative results of object grounding improvement between HOV-SG~\cite{werby2024hierarchical} and OVIGo-3DHSG on our complex object queries benchmark on HM3DSem~\cite{yadav2023habitat}.}
    \label{tab:habitat_objects}
    \begin{tabularx}{\textwidth}{
    |>{\centering\arraybackslash}p{3.2cm}
    |>{\centering\arraybackslash}p{2.9cm}
    |>{\centering\arraybackslash}p{2.9cm}
    |>{\centering\arraybackslash}p{2.9cm}
    |>{\centering\arraybackslash}p{2.9cm}|
    }
        \toprule 
        Metric & $Accuracy@0.1$ & $Accuracy@0.25$ & $Accuracy@0.5$ & $Accuracy@0.75$\\
        \midrule
        $\Delta$ (Improvement) & 15.62 & 12.54 & 9.34 & 5.21 \\
        \bottomrule
    \end{tabularx}
\end{table}

\section{Conclusion}

Experimental results on the HM3DSem dataset demonstrates the effectiveness of OVIGo-3DHSG method, that builds a 3D hierarchical scene graph to represent complex indoor environments, achieving high semantic and geometric accuracy and outperforming existing methods in object grounding tasks. The results indicate that OVIGo-3DHSG is a promising solution for advancing indoor multifloor spatial reasoning and grounding through natural language.

{
\renewcommand{\mkbibbrackets}[1]{#1}
\DeclareFieldFormat{labelnumber}{#1.}
\printbibliography

@inproceedings{yue2023connecting,
  title     = {{Connecting the Dots: Floorplan Reconstruction Using Two-Level Queries}},
  author    = {Yue, Yuanwen and Kontogianni, Theodora and Schindler, Konrad and Engelmann, Francis},
  booktitle = {IEEE/CVF Conference on Computer Vision and Pattern Recognition (CVPR)},
  year      = {2023}
}

@software{yolo11_ultralytics,
  author = {Glenn Jocher and Jing Qiu},
  title = {Ultralytics YOLO11},
  version = {11.0.0},
  year = {2024},
  url = {https://github.com/ultralytics/ultralytics},
  orcid = {0000-0001-5950-6979, 0000-0002-7603-6750, 0000-0003-3783-7069},
  license = {AGPL-3.0}
}

@inproceedings{yeshwanthliu2023scannetpp,
  title={ScanNet++: A High-Fidelity Dataset of 3D Indoor Scenes},
  author={Yeshwanth, Chandan and Liu, Yueh-Cheng and Nie{\ss}ner, Matthias and Dai, Angela},
  booktitle = {Proceedings of the International Conference on Computer Vision ({ICCV})},
  year={2023}
}

@inproceedings{Structured3D,
  title     = {Structured3D: A Large Photo-realistic Dataset for Structured 3D Modeling},
  author    = {Jia Zheng and Junfei Zhang and Jing Li and Rui Tang and Shenghua Gao and Zihan Zhou},
  booktitle = {Proceedings of The European Conference on Computer Vision (ECCV)},
  year      = {2020}
}

@inproceedings{DBSCAN,
author = {Ester, Martin and Kriegel, Hans-Peter and Sander, J\"{o}rg and Xu, Xiaowei},
title = {A density-based algorithm for discovering clusters in large spatial databases with noise},
year = {1996},
publisher = {AAAI Press},
booktitle = {Proceedings of the Second International Conference on Knowledge Discovery and Data Mining},
pages = {226–231},
numpages = {6},
location = {Portland, Oregon},
series = {KDD'96}
}

@ARTICLE{Otsu,
  author={Otsu, Nobuyuki},
  journal={IEEE Transactions on Systems, Man, and Cybernetics}, 
  title={A Threshold Selection Method from Gray-Level Histograms}, 
  year={1979},
  volume={9},
  number={1},
  pages={62-66},
  doi={10.1109/TSMC.1979.4310076}}

@article{mildenhall2021nerf,
  title={Nerf: Representing scenes as neural radiance fields for view synthesis},
  author={Mildenhall, Ben and Srinivasan, Pratul P and Tancik, Matthew and Barron, Jonathan T and Ramamoorthi, Ravi and Ng, Ren},
  journal={Communications of the ACM},
  volume={65},
  number={1},
  pages={99--106},
  year={2021},
  publisher={ACM New York, NY, USA}
}

@article{kerbl20233d,
  title={3d gaussian splatting for real-time radiance field rendering.},
  author={Kerbl, Bernhard and Kopanas, Georgios and Leimk{\"u}hler, Thomas and Drettakis, George},
  journal={ACM Trans. Graph.},
  volume={42},
  number={4},
  pages={139--1},
  year={2023}
}

@article{macario2022comprehensive,
  title={A comprehensive survey of visual slam algorithms},
  author={Macario Barros, Andr{\'e}a and Michel, Maugan and Moline, Yoann and Corre, Gwenol{\'e} and Carrel, Fr{\'e}d{\'e}rick},
  journal={Robotics},
  volume={11},
  number={1},
  pages={24},
  year={2022},
  publisher={MDPI}
}

@article{lee2024comparative,
  title={Comparative Analysis of YOLO Series (from V1 to V11) and Their Application in Computer Vision},
  author={Lee, Yong-Hwan and Kim, Heung-Jun},
  journal={Journal of the Semiconductor \& Display Technology},
  volume={23},
  number={4},
  pages={190--198},
  year={2024},
  publisher={The Korean Society Of Semiconductor \& Display Technology}
}

@article{minaee2021image,
  title={Image segmentation using deep learning: A survey},
  author={Minaee, Shervin and Boykov, Yuri and Porikli, Fatih and Plaza, Antonio and Kehtarnavaz, Nasser and Terzopoulos, Demetri},
  journal={IEEE transactions on pattern analysis and machine intelligence},
  volume={44},
  number={7},
  pages={3523--3542},
  year={2021},
  publisher={IEEE}
}

@article{zou2023segment,
  title={Segment everything everywhere all at once},
  author={Zou, Xueyan and Yang, Jianwei and Zhang, Hao and Li, Feng and Li, Linjie and Wang, Jianfeng and Wang, Lijuan and Gao, Jianfeng and Lee, Yong Jae},
  journal={Advances in neural information processing systems},
  volume={36},
  pages={19769--19782},
  year={2023}
}

@inproceedings{liu2024grounding,
  title={Grounding dino: Marrying dino with grounded pre-training for open-set object detection},
  author={Liu, Shilong and Zeng, Zhaoyang and Ren, Tianhe and Li, Feng and Zhang, Hao and Yang, Jie and Jiang, Qing and Li, Chunyuan and Yang, Jianwei and Su, Hang and others},
  booktitle={European Conference on Computer Vision},
  pages={38--55},
  year={2024},
  organization={Springer}
}

@inproceedings{cheng2024yolo,
  title={Yolo-world: Real-time open-vocabulary object detection},
  author={Cheng, Tianheng and Song, Lin and Ge, Yixiao and Liu, Wenyu and Wang, Xinggang and Shan, Ying},
  booktitle={Proceedings of the IEEE/CVF Conference on Computer Vision and Pattern Recognition},
  pages={16901--16911},
  year={2024}
}

@article{sun2024sparse,
  title={Sparse Voxels Rasterization: Real-time High-fidelity Radiance Field Rendering},
  author={Sun, Cheng and Choe, Jaesung and Loop, Charles and Ma, Wei-Chiu and Wang, Yu-Chiang Frank},
  journal={arXiv preprint arXiv:2412.04459},
  year={2024}
}

@inproceedings{zou20253d,
  title={3D-SPATIAL MULTIMODAL MEMORY},
  author={Zou, Xueyan and Song, Yuchen and Qiu, Ri-Zhao and Peng, Xuanbin and Ye, Jianglong and Liu, Sifei and Wang, Xiaolong},
  booktitle={The Thirteenth International Conference on Learning Representations},
  year={2025}
}

@article{yu2025rgb,
  title={Rgb-only gaussian splatting slam for unbounded outdoor scenes},
  author={Yu, Sicheng and Cheng, Chong and Zhou, Yifan and Yang, Xiaojun and Wang, Hao},
  journal={arXiv preprint arXiv:2502.15633},
  year={2025}
}

@inproceedings{armeni20193d,
  title={3d scene graph: A structure for unified semantics, 3d space, and camera},
  author={Armeni, Iro and He, Zhi-Yang and Gwak, JunYoung and Zamir, Amir R and Fischer, Martin and Malik, Jitendra and Savarese, Silvio},
  booktitle={Proceedings of the IEEE/CVF international conference on computer vision},
  pages={5664--5673},
  year={2019}
}

@article{rosinol20203d,
  title={3D dynamic scene graphs: Actionable spatial perception with places, objects, and humans},
  author={Rosinol, Antoni and Gupta, Arjun and Abate, Marcus and Shi, Jingnan and Carlone, Luca},
  journal={arXiv preprint arXiv:2002.06289},
  year={2020}
}

@article{hughes2022hydra,
  title={Hydra: A real-time spatial perception system for 3D scene graph construction and optimization},
  author={Hughes, Nathan and Chang, Yun and Carlone, Luca},
  journal={arXiv preprint arXiv:2201.13360},
  year={2022}
}

@article{bavle2023s,
  title={S-graphs+: Real-time localization and mapping leveraging hierarchical representations},
  author={Bavle, Hriday and Sanchez-Lopez, Jose Luis and Shaheer, Muhammad and Civera, Javier and Voos, Holger},
  journal={IEEE Robotics and Automation Letters},
  volume={8},
  number={8},
  pages={4927--4934},
  year={2023},
  publisher={IEEE}
}

@article{hughes2024foundations,
  title={Foundations of spatial perception for robotics: Hierarchical representations and real-time systems},
  author={Hughes, Nathan and Chang, Yun and Hu, Siyi and Talak, Rajat and Abdulhai, Rumaia and Strader, Jared and Carlone, Luca},
  journal={The International Journal of Robotics Research},
  volume={43},
  number={10},
  pages={1457--1505},
  year={2024},
  publisher={SAGE Publications Sage UK: London, England}
}

@inproceedings{werby2024hierarchical,
  title={Hierarchical open-vocabulary 3d scene graphs for language-grounded robot navigation},
  author={Werby, Abdelrhman and Huang, Chenguang and B{\"u}chner, Martin and Valada, Abhinav and Burgard, Wolfram},
  booktitle={First Workshop on Vision-Language Models for Navigation and Manipulation at ICRA 2024},
  year={2024}
}

@article{tang2024openobject,
  title={OpenObject-NAV: Open-Vocabulary Object-Oriented Navigation Based on Dynamic Carrier-Relationship Scene Graph},
  author={Tang, Yujie and Wang, Meiling and Deng, Yinan and Zheng, Zibo and Zhong, Jiagui and Yue, Yufeng},
  journal={arXiv preprint arXiv:2409.18743},
  year={2024}
}

@article{maggio2024clio,
  title={Clio: Real-time task-driven open-set 3d scene graphs},
  author={Maggio, Dominic and Chang, Yun and Hughes, Nathan and Trang, Matthew and Griffith, Dan and Dougherty, Carlyn and Cristofalo, Eric and Schmid, Lukas and Carlone, Luca},
  journal={IEEE Robotics and Automation Letters},
  year={2024},
  publisher={IEEE}
}

@article{linok2024beyond,
  title={Beyond Bare Queries: Open-Vocabulary Object Grounding with 3D Scene Graph},
  author={Linok, Sergey and Zemskova, Tatiana and Ladanova, Svetlana and Titkov, Roman and Yudin, Dmitry and Monastyrny, Maxim and Valenkov, Aleksei},
  journal={arXiv preprint arXiv:2406.07113},
  year={2024}
}

@inproceedings{yuan2024visual,
  title={Visual programming for zero-shot open-vocabulary 3d visual grounding},
  author={Yuan, Zhihao and Ren, Jinke and Feng, Chun-Mei and Zhao, Hengshuang and Cui, Shuguang and Li, Zhen},
  booktitle={Proceedings of the IEEE/CVF Conference on Computer Vision and Pattern Recognition},
  pages={20623--20633},
  year={2024}
}

@inproceedings{gu2024conceptgraphs,
  title={Conceptgraphs: Open-vocabulary 3d scene graphs for perception and planning},
  author={Gu, Qiao and Kuwajerwala, Ali and Morin, Sacha and Jatavallabhula, Krishna Murthy and Sen, Bipasha and Agarwal, Aditya and Rivera, Corban and Paul, William and Ellis, Kirsty and Chellappa, Rama and others},
  booktitle={2024 IEEE International Conference on Robotics and Automation (ICRA)},
  pages={5021--5028},
  year={2024},
  organization={IEEE}
}

@inproceedings{jia2024sceneverse,
  title={Sceneverse: Scaling 3d vision-language learning for grounded scene understanding},
  author={Jia, Baoxiong and Chen, Yixin and Yu, Huangyue and Wang, Yan and Niu, Xuesong and Liu, Tengyu and Li, Qing and Huang, Siyuan},
  booktitle={European Conference on Computer Vision},
  pages={289--310},
  year={2024},
  organization={Springer}
}

@article{kornilov2018overview,
  title={An overview of watershed algorithm implementations in open source libraries},
  author={Kornilov, Anton S and Safonov, Ilia V},
  journal={Journal of Imaging},
  volume={4},
  number={10},
  pages={123},
  year={2018},
  publisher={MDPI}
}

@article{zhang2023recognize,
  title={Recognize Anything: A Strong Image Tagging Model},
  author={Zhang, Youcai and Huang, Xinyu and Ma, Jinyu and Li, Zhaoyang and Luo, Zhaochuan and Xie, Yanchun and Qin, Yuzhuo and Luo, Tong and Li, Yaqian and Liu, Shilong and others},
  journal={arXiv preprint arXiv:2306.03514},
  year={2023}
}

@article{zhang2023mobilesamv2,
  title={Mobilesamv2: Faster segment anything to everything},
  author={Zhang, Chaoning and Han, Dongshen and Zheng, Sheng and Choi, Jinwoo and Kim, Tae-Ho and Hong, Choong Seon},
  journal={arXiv preprint arXiv:2312.09579},
  year={2023}
}

@article{wei2022chain,
  title={Chain-of-thought prompting elicits reasoning in large language models},
  author={Wei, Jason and Wang, Xuezhi and Schuurmans, Dale and Bosma, Maarten and Xia, Fei and Chi, Ed and Le, Quoc V and Zhou, Denny and others},
  journal={Advances in neural information processing systems},
  volume={35},
  pages={24824--24837},
  year={2022}
}

@inproceedings{yadav2023habitat,
  title={Habitat-matterport 3d semantics dataset},
  author={Yadav, Karmesh and Ramrakhya, Ram and Ramakrishnan, Santhosh Kumar and Gervet, Theo and Turner, John and Gokaslan, Aaron and Maestre, Noah and Chang, Angel Xuan and Batra, Dhruv and Savva, Manolis and others},
  booktitle={Proceedings of the IEEE/CVF Conference on Computer Vision and Pattern Recognition},
  pages={4927--4936},
  year={2023}
}

@article{khanam2024yolov11,
  title={Yolov11: An overview of the key architectural enhancements},
  author={Khanam, Rahima and Hussain, Muhammad},
  journal={arXiv preprint arXiv:2410.17725},
  year={2024}
}

@inproceedings{huang2023visual,
  title={Visual language maps for robot navigation},
  author={Huang, Chenguang and Mees, Oier and Zeng, Andy and Burgard, Wolfram},
  booktitle={2023 IEEE International Conference on Robotics and Automation (ICRA)},
  pages={10608--10615},
  year={2023},
  organization={IEEE}
}

@inproceedings{yudin2024multimodal,
  title={Multimodal 3D map reconstruction for intelligent robotcs using neural network-based methods},
  author={Yudin, Dmitrii Aleksandrovich},
  booktitle={Doklady Mathematics},
  volume={110},
  number={Suppl 1},
  pages={S117--S125},
  year={2024},
  organization={Springer}
}

@article{mironov2023strl,
  title={STRL-Robotics: intelligent control for robotic platform in human-oriented environment},
  author={Mironov, Konstantin V and Yudin, Dmitry A and Alhaddad, Muhammad and Makarov, Dmitry A and Pushkarev, Daniil S and Linok, Sergey A and Belkin, Ilya V and Krishtopik, Andrey S and Golovin, Vladislav A and Yakovlev, Konstantin S and others},
  journal={Artificial Intelligence and Decision Making},
  number={2},
  pages={45--63},
  year={2023}
}

@article{linok2023influence,
  title={Influence of neural network receptive field on monocular depth and ego-motion estimation},
  author={Linok, SA and Yudin, DA},
  journal={Optical Memory and Neural Networks},
  volume={32},
  number={Suppl 2},
  pages={S206--S213},
  year={2023},
  publisher={Springer}
}

@article{zemskova20243dgraphllm,
  title={3DGraphLLM: Combining Semantic Graphs and Large Language Models for 3D Scene Understanding},
  author={Zemskova, Tatiana and Yudin, Dmitry},
  journal={arXiv preprint arXiv:2412.18450},
  year={2024}
}

@article{hurst2024gpt,
  title={Gpt-4o system card},
  author={Hurst, Aaron and Lerer, Adam and Goucher, Adam P and Perelman, Adam and Ramesh, Aditya and Clark, Aidan and Ostrow, AJ and Welihinda, Akila and Hayes, Alan and Radford, Alec and others},
  journal={arXiv preprint arXiv:2410.21276},
  year={2024}
}
}
\end{document}